\documentclass[preprint,journal]{vgtc}       

\ifpdf
  \pdfoutput=1\relax                   
  \usepackage{graphicx}                
  \DeclareGraphicsExtensions{.pdf,.png,.jpg,.jpeg} 
\else
  \usepackage{graphicx}                
  \DeclareGraphicsExtensions{.eps}     
\fi%

\graphicspath{{figures/}{pictures/}{images/}{./}} 
\usepackage[switch]{lineno} 
\usepackage{microtype}                 
\PassOptionsToPackage{warn}{textcomp}  
\usepackage{textcomp}                  
\usepackage{mathptmx}                  
\usepackage{times}                     
\usepackage{cite}                      
\usepackage{tabu}                      
\usepackage{booktabs}                  
\usepackage{algorithmicx}
\usepackage[ruled]{algorithm}
\usepackage[noend]{algpseudocode}
\usepackage{amsmath}
\usepackage{amssymb}

\usepackage{subfig}

\newcommand*{\img}[1]{%
    \raisebox{0\baselineskip}{%
        \includegraphics[
        height=0.6\baselineskip,
        width=0.6\baselineskip,
        keepaspectratio,
        ]{#1}%
    }%
}

\usepackage{xcolor}

\newcommand{\wm}{\textcolor{black}}

\ieeedoi{10.1109/TVCG.2024.3471181}

\onlineid{1627}

\vgtccategory{Research}
\vgtcpapertype{algorithm/technique}

\title{HUMAP: Hierarchical Uniform Manifold Approximation and Projection}

\author{Wilson E. Marcílio-Jr, Danilo M. Eler, Fernando V. Paulovich, Member, IEEE, Rafael M. Martins, Member, IEEE}
\authorfooter{

\item
 W. E. Marcílio-Jr is with São Paulo State University (UNESP) and Nomic AI, Brazil. E-mails: wilson.marcilio@unesp.br, wilson@nomic.ai.

\item
 D. M. Eler is with São Paulo State University (UNESP), Brazil. E-mail: danilo.eler@unesp.br.
 
 \item
  F. V. Paulovich is with Eindhoven University of Technology, the Netherlands. E-mail: f.paulovich@tue.nl
   
  \item
  R. M. Martins is with Linnaeus University, Vaxjö, Sweden E-mail: rafael.martins@lnu.se
}

\shortauthortitle{Marcílio-Jr \MakeLowercase{\textit{et al.}}: HUMAP: Hierarchical Uniform Manifold Approximation and Projection}

\abstract{
Dimensionality reduction (DR) techniques help analysts to understand patterns in high-dimensional spaces. These techniques, often represented by scatter plots, are employed in diverse science domains and facilitate similarity analysis among clusters and data samples. For datasets containing many granularities or when analysis follows the information visualization mantra, hierarchical DR techniques are the most suitable approach since they present major structures beforehand and details on demand. 
This work presents HUMAP, a novel hierarchical dimensionality reduction technique designed to be flexible on preserving local and global structures and preserve the mental map throughout hierarchical exploration. We provide empirical evidence of our technique's superiority compared with current hierarchical approaches and show a case study applying HUMAP for dataset labelling.

} 


\CCScatlist{ 
 \CCScat{K.6.1}{Management of Computing and Information Systems}%
{Project and People Management}{Life Cycle};
 \CCScat{K.7.m}{The Computing Profession}{Miscellaneous}{Ethics}
}

\teaser{
 \centering
 \includegraphics[width=\linewidth]{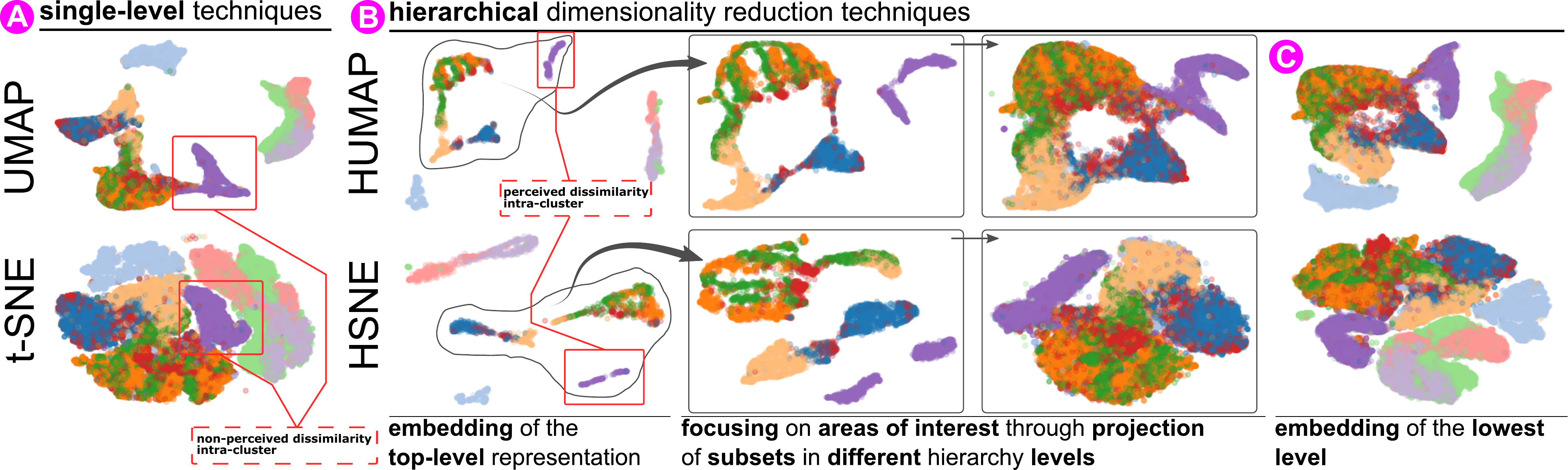}
 \caption{Comparison of traditional dimensionality reduction (DR) (A) vs. hierarchical DR (HDR) techniques (B). Traditional techniques operate in a single level of detail and focus on delivering a big picture that depicts the dataset as a whole, ignoring fine-grained details of intra-cluster distributions. On the other hand, HDR approaches summarize important intra-cluster information on the top-level representations of the hierarchy imposed on the input data without affecting the layout’s overall organization. In a HUMAP projection, lower hierarchy levels resemble superior hierarchy levels, and users receive more information about the similarity among clusters when more detail (more data points) is required (B,C). 
 }
 \label{fig:fmnist-teaser}
}

\vgtcinsertpkg

\begin{document}


\firstsection{Introduction}

\maketitle

The majority of knowledge domains now require high-dimensional data analysis as a fundamental task. 
Dimension reduction (DR) is one of the most notable examples of how visualization tools for analysis and navigation are becoming more and more crucial in these situations. DR techniques generally aim to decrease the dimensionality of a dataset while maintaining the structure present in the initial high-dimensional space~\cite{Nonato2019}. The analysis of document collections~\cite{Eler2009}, gene expression data~\cite{Unen2017, Salamon2018, Hollt2018, Somarakis2019, Krueger2020}, and features learned by deep learning models~\cite{Rauber2017, Benato2018} are just a few tasks supported by DR results (or embeddings), which are represented as scatter plots. Practitioners typically search for DR techniques for these applications that can reveal complex structures (like manifolds), \wm{encode clusters and their relationship.}

Traditionally, DR techniques operate at a single level of detail, resulting in layouts that emphasize general patterns that convey the overall distribution of points in the high-dimensional space. 
\wm{The layout generated by these techniques often hide important dissimilarity within clusters.}
To solve this problem, one could always create a new projection using a cluster of interest or a subset of data instances, but the resulting layout would not take into account the contribution of the other points, and context would not be preserved. Another solution is to employ hierarchical DR (HDR) techniques, which encourage the exploration of DR outcomes using the maxim of visual information-seeking \wm{mantra}~\cite{Shneiderman1996}: \emph{Overview first, zoom and filter, then details-on-demand}.


Although groups inside clusters are better encoded in HDR, the detailed structures seen in conventional representations are summarized. By examining lower (and more detailed) hierarchy levels through expanding clusters in HDR representations, users can learn more about the relationships among the subclusters that are more visible on the top-level representation. This concept is well-illustrated for the class in purple on Fig.~\ref{fig:fmnist-teaser}, highlighted by a red box. Traditional approaches do not allow for the top-level representations to show the separation of two dominant structures (the purple cluster highlighted by a red box in Fig.~\ref{fig:fmnist-teaser}(B)).

The preservation of mental maps is another component of our strategy for preserving context during hierarchical exploration. This is crucial because the data undergoes numerous non-linear transformations following an embedding step, as shown for HSNE in Fig.~\ref{fig:fmnist-teaser}(B) for the second and third embeddings. Notice how the positioning and structural relationships between clusters are lost during layout optimization, necessitating the maintenance of the mental map---particularly for unlabeled datasets.


The available methods for hierarchical dimensionality reduction present a few issues, mainly related to scale~\cite{Paulovich2008Hipp, Kuchroo2020}, ability to uncover complex structures~\cite{Kuchroo2020}, and mental map preservation~\cite{Pezzotti2016}. 
The preservation of the mental map across hierarchy levels and maintaining a good trade-off between the preservation of global and local structures in the low-dimensional representation of the dataset are two HDR design considerations that these techniques cannot fully address. 

To meet these challenges, we present Hierarchical Uniform Manifold Approximation and Projection (HUMAP). Our HDR technique, which is based on UMAP~\cite{McInnes2018}, creates a hierarchy on the dataset by encoding both global and local similarity information between data points in a two-step similarity definition. HUMAP successfully reveals complex structures present in the numerous granularities of a dataset while maintaining the mental map as the user drills down the hierarchy by using projected data points from higher levels to influence the low-dimensional representation of lower hierarchy levels. We provide experimental evidence that HUMAP addresses the aforementioned design considerations through visual and quantitative evaluation. We also offer an in-depth case study of real-world data. In summary, the contribution of this paper is a novel HDR technique that improves the representation of both global and local relations while preserving the mental map across hierarchy levels. Our code is available on GitHub\footnote{https://github.com/wilsonjr/humap}.

The remainder of this paper is organized as follows. Section~\ref{sec:related-works} presents the related works. We describe our technique in Section~\ref{sec:humap} and present a case study in Section~\ref{sec:case-studies}. In Section~\ref{sec:experiments} we quantitatively evaluate HUMAP and compare it against other HDR techniques. Discussions are presented in Section~\ref{sec:discussion} and in Section~\ref{sec:conclusion} we draw our conclusions.
\section{Related Work}
\label{sec:related-works}

By reducing the number of dimensions while maintaining structural and similarity relations in the low-dimensional representation, DR techniques aid in the analysis of high-dimensional datasets~\cite{Nonato2019}. Users typically search for patterns in the data when using a DR technique, such as clusters, shapes, and outliers. To ensure a successful data exploration process, it is essential to understand the data and the available DR techniques. For instance, linear DR techniques are well known for quickly revealing variability~\cite{Jolliffe1986, Paulovich2008, Joia2011}. While generally being more difficult and time-consuming, non-linear DR techniques~\cite{Maaten2008, Pezzotti2016, McInnes2018, Moon2019} can reveal more complex structures that are present in high-dimensional space.


The scatterplot metaphor is frequently employed represent DR results. Despite being widely used for exploratory data analysis~\cite{Micallef2017}, scatter plots lack effectiveness due to marker overlap~\cite{Sedlmair2013, Sarikaya2018}. When exploring large datasets as a whole, traditional DR techniques may obscure crucial details within and between clusters. For instance, some datasets exhibit inherently multilevel structures~\cite{Lahnemann2020}, which call for a DR technique to be more adaptable. Knowledge discovery is facilitated in this way by strategies that make exploration simpler and direct users through the exploratory process. One such strategy is the use of hierarchical dimensionality reduction (HDR) techniques, which offer exploration mechanisms based on the mantra \textit{overview first \& details-on-demand}~\cite{Shneiderman1996}, concentrating on essential information as needed by the user.


There are many traditional DR strategies (see~\cite{Nonato2019, Espadoto2019} for helpful surveys), but very few hierarchical DR (HDR) methods~\cite{Nonato2019, Hollt2019}. These HDR methods have the feature of first establishing the hierarchical structure before enabling multilevel exploration. The challenge is conveying different levels while preserving context and neighborhood structures~\cite{Nonato2019}. Therefore, current studies concentrate on defining the hierarchical structure while using the projection engine of previous traditional DR approaches. For example, in order to solve the MDS computational complexity problem, MDSteer~\cite{Williams2004} incrementally computes a multidimensional scaling layout in response to user demand. Glimmer~\cite{Ingram2009} also addresses MDS complexity by performing projection using a multilevel GPU scheme. Simply put, the authors interpolate high hierarchy levels to create a hierarchy. However, MDS-based approaches do not adequately capture the complex structures (such as manifolds) found in the majority of real-world and practical datasets (e.g., deep learning features, image collections, or biological data). Hierarchical PCA variants suffer from a lack of non-linear structure communication strategies~\cite{Westerhuis1998, Janne2001, Agarwal2007}.

%
%

Data organization in HiPP~\cite{Paulovich2008Hipp} involves landmarks and hierarchical clustering. In finer levels, the data points are represented and influenced by the landmarks of coarser levels. HiPP uses a force algorithm to deal with overlaps when positioning points heuristically and takes into account the landmarks of LSP~\cite{Paulovich2008} for context preservation and to communicate hierarchy levels. One of the most reliable methods is HSNE~\cite{Pezzotti2016}. The HSNE algorithm builds a hierarchy to preserve global and local relationships at high hierarchy levels using random walks on a transition matrix. The analysis is hampered when mental map preservation is crucial because the HSNE's embeddings lose the structural relationships that were presented at higher levels during hierarchical exploration. To provide interactive analysis in a reasonable amount of time, HSNE also needs a GPU. Using diffusion condensation~\cite{Brugnone2019}, a recent method known as Multiscale PHATE~\cite{Kuchroo2020} computes a manifold-intrinsic diffusion space on the input data and condenses data points towards centroids to produce groups of multiple granularities. Massive datasets can be handled by multiscale PHATE, but only for a few dimensions. Additionally, it appears to have issues with its embedding engine when the input dataset does not contain continuous phenomena~\cite{Moon2018}. The hierarchy levels of these methods all share the inclusion of representative samples or landmarks. It is interesting to note that any single-level-of-detail dimensionality reduction method that projects data points onto landmarks is capable of having a hierarchical version~\cite{Nonato2019}.

\begin{figure*}
 \centering 
 \includegraphics[width=\linewidth]{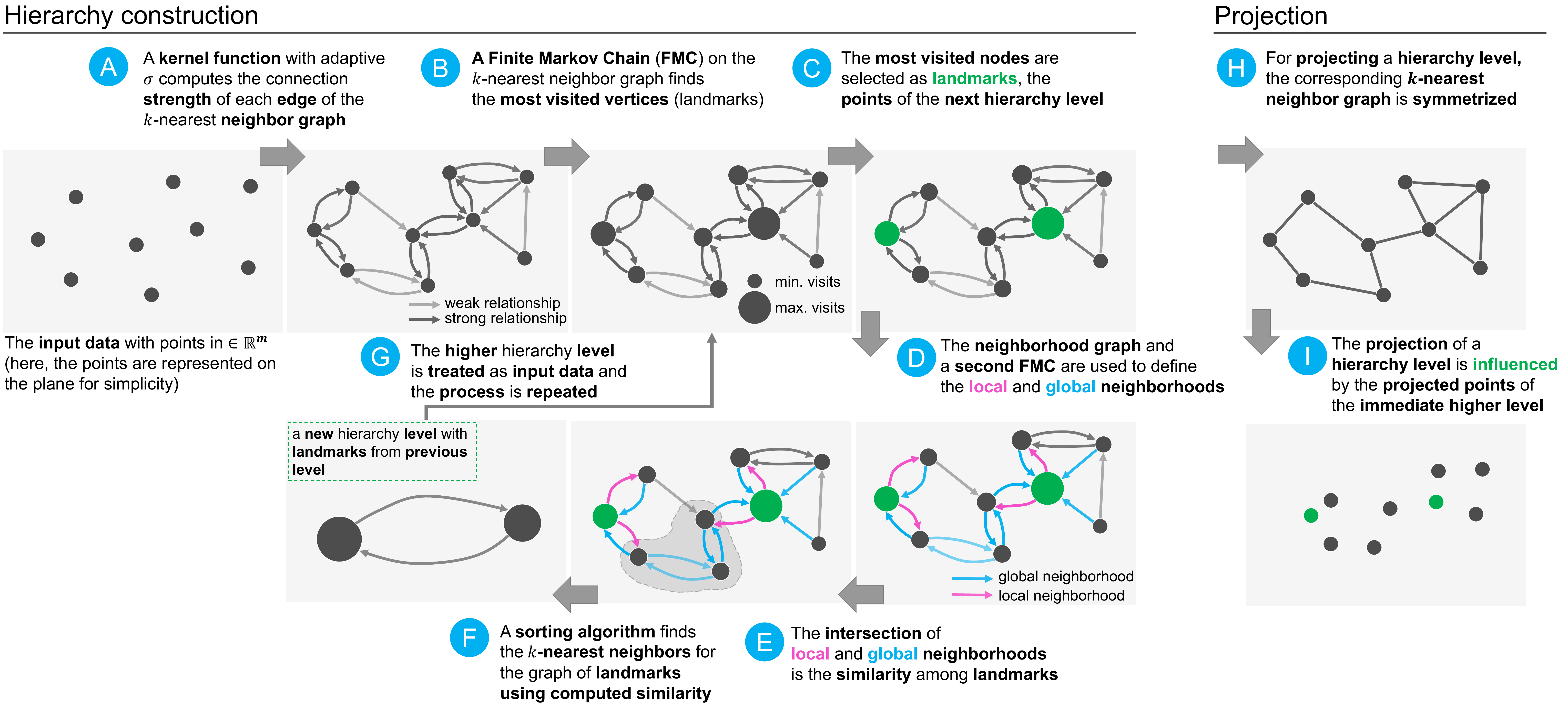}
 \caption{The hierarchy is built from the bottom up. First, the connection strength between data points in the high-dimensional space is determined using a k-nearest neighbor graph and a kernel function (A). After several random walk steps, the graph's structure and the strength of the connections are used to determine which nodes have been visited the most (B). The landmarks correspond to the points in the higher hierarchy level (C). To repeat the same procedure for high hierarchy levels, we calculate the intersection of representation neighborhoods (E), which were formed by joining local and global neighborhoods (D). With the exception of the first hierarchy level (whole dataset), we compute the k-nearest neighbors using a sorting algorithm (F). We employ a modified UMAP~\cite{McInnes2018} optimization for projecting hierarchy levels (or subsets of them). \wm{Finally, the graph is symmetrized (H) and coordinates of projected points influence the positioning subsequent levels (I)}.}
 \label{fig:overview-humap}
\end{figure*}

Finally, users work with embeddings of increasing size during hierarchical exploration. Thus, it is essential to maintain the mental map between successive projections in order to prevent users from being tricked by geometrical transformations during exploratory analysis. \wm{This task, often refered as \textit{projection alignment}, can be addresses in different ways.} A time-varying dataset is projected using dynamic t-SNE~\cite{Rauber2016} to reduce temporal variability that is unimportant to the final layout. This is accomplished by including a term in the cost equation for the t-SNE that regulates the trade-off between alignment and conventional t-SNE optimization. A trade-off parameter is also used in VFF~\cite{Hilasaca2020} to regulate the alignment of projections in feature fusion tasks. VFF, however, is ineffective for preserving local structures~\cite{Cantareira2020}. Finally, Cantareira and Paulovich~\cite{Cantareira2020} proposed a general model for projection alignment that incorporates the original cost function of the DR technique and a penalty term applied for alignment. \wm{Another important problem that these techniques address is when dealing with streaming applications, where the use of DR techniques for conveying information has grown~\cite{MOHEDANOMUNOZ2023120252}. The preservation of mental maps is critical in such scenarios, as users would have to expend a significant amount of cognitive effort tracking the position of data points in the visual space.}. Our approach to maintaining the mental map entails directing the movement of data samples that have already been projected at high hierarchy levels and affecting the placement of new data samples. Fig.~\ref{fig:fmnist-teaser} illustrates this. The highest hierarchy level (top-level) of HUMAP resembles the lowest hierarchy level and maintains the structures of chosen subsets (B). In contrast to single-level techniques (A), it summarizes the data while providing information on the connections between and within clusters.

\section{Hierarchical Uniform Manifold Approximation and Projection (HUMAP)}\label{sec:humap}

Two main components of our HDR technique are \textit{Hierarchy construction} and \textit{Projection} (Fig.~\ref{fig:overview-humap}). In the former, we use a similarity measure between landmarks to create a tree-like structure on the high-dimensional dataset. In the latter, we incorporate the hierarchy levels in response to the user's demand for more specific data. All the steps (A–G) in this process are shown in Fig.~\ref{fig:overview-humap}.

The first step in building a hierarchy from bottom to top is to use a kernel function to determine the connection strengths (local affinities) of a $k$-nearest neighbor graph of data points in the high-dimensional space $\mathbb{R}^m$ (step A). Then, similar to previous research~\cite{Pezzotti2016}, we employ the Finite Markov Chain (FMC) to identify the most visited nodes, which consist of the landmarks for the higher level (step C). Steps (D) and (E) of the FMC process are used to encode local and global neighborhood information for each landmark as well as to build a neighborhood structure for high hierarchy levels ($>1$). In order to define a new hierarchy level, a new neighborhood graph is created in step (F) using the computed similarity (step G).

For projecting hierarchy levels, the neighborhood graph is first symmetrized (step H), so each edge's strength helps in finding a suitable position in the low-dimensional representation. For the purpose of maintaining mental maps, the projection of lower levels, with the exception of the top hierarchy level, is influenced by the low-dimensional positions (I) of data points in higher levels, which we discuss in Section~\ref{sec:mental-map}.

We use a few strategies from UMAP~\cite{McInnes2018} to build our approach, including the kernel function to calculate the connection strengths among data samples and the embedding strategy and concepts from HSNE~\cite{Pezzotti2016} for landmark selection. We concentrate on the HUMAP components for hierarchical dimensionality reduction in the following section.

\subsection{Landmark selection}

DR methods frequently use $k$-nearest neighbor graphs to capture manifolds in high-dimensional spaces. The path formed by the vertices in the $k$-nearest neighbor graph is used to represent the distance between two arbitrary points in the high-dimensional space in this process, which approximates global distances by adding up local relations~\cite{Moon2019}. Therefore, the first step in many DR techniques that can find manifolds in data is to fit a $k$-nearest neighbor graph~\cite{Maaten2008, McInnes2018, Moon2019}. 

After finding the $k$-nearest neighbor graph, a kernel function defines the strength (or probability) of every edge in the graph. Here, we adopted UMAP's~\cite{McInnes2018} kernel function for two data points $x_i, x_j \in X$ (the high-dimensional dataset):

\begin{equation}
    \label{eq:kernel}
p_{i|j} = e^{-\frac{d(x_i, x_j) -\rho_i}{\sigma_i}}
\end{equation}

where $d(x_i, x_j)$ is the euclidean distance between the data points $x_i$ and $x_j$, $\rho_i$ is the euclidean distance of $x_i$ to its closest neighbor, and $\sigma_i$ is a kernel value that depends on the number of neighbors $k$, according to the following equation~\cite{McInnes2018},

\begin{equation}
    \label{eq:k}
    k = 2^{\sum\limits_i p_{i|j}}.
\end{equation}

Notice that $\sigma_i$ is unknown. However, by fixing a value for $k$, we can use Equation~\ref{eq:k} to compute it using a binary search~\cite{McInnes2018}. That is, for each data point $x_i$ (matrix row), Equation~\ref{eq:kernel} is plugged into Equation~\ref{eq:k} and a binary search is used to find the best $\sigma_i$ value given the number of nearest neighbors ($k$). Thus, the value for $\sigma_i$ that produces a $k'$ closest to the actual $k$ is selected. This procedure ensures a different kernel value ($\sigma_i$) for each data point and its neighborhood. For a fixed $k$, higher $\sigma_i$ values encode dense neighborhoods while lower $\sigma_i$ values encode sparse neighborhoods. In other words, $\sigma_i$ encodes how close the data points are in the neighborhood of $x_i$. Lastly, the parameter $\rho_i$ gives a locally adaptive kernel for each data point and ensures a good topological representation of high-dimensional data~\cite{McInnes2018}.

The connection strength between each neighbor is determined by computing $p_{i|j}$ for each pair of data points in accordance with the neighborhood density (please, see Equation~\ref{eq:kernel}). The dataset is summarized at higher hierarchical levels using good representative data points (or landmarks) based on the strengths of these connections. For this task, we use random walks on a finite markov chain, as in Pezzotti et al.'s~\cite{Pezzotti2016} work (FMC). The path from an initial state to the final state after $\mu$ steps (or hops) on the neighborhood graph is known as a random walk and is defined by a length. As a result, we compute the following equation to generate the transition probabilities for the FMC from $p_{i|j}$:

\begin{equation}
    \label{eq:transition}
    T_{i, j} = \frac{p_{i|j}}{\sum\limits_{k}p_{i|k}}\text{  }(j, k \in NH(i)),
\end{equation}

where data points $x_j$ and $x_k$ are in the neighborhood of $x_i$ ($NH(i)$). Then, the landmarks are sampled from $T_{i,j}$ by a simple Markov Chain Monte Carlo technique~\cite{Gey2011}. For each data point $i$, we start $n$ random walks with length $\mu$. We empirically evaluated that $n = \mu = 10$ results in embeddings that trustfully represent the underlying data structure. This whole process, from kernel smoothing to landmark selection, is represented by the steps from (A) to (C) in Fig.~\ref{fig:overview-humap}.

Unlike previous approaches~\cite{Pezzotti2016, Kuchroo2020}, users have control over the number of data points in each hierarchy level, making it easier to experiment with datasets with varying characteristics (e.g., cluster density)---though other threshold-based approaches can be used~\cite{Pezzotti2016}. By specifying the number of landmarks on level $i$ ($|H^i|$), users control the number of data points in a hierarchy level. We store the random walk endpoints and choose the $|H^i|$ most frequented data points as landmarks during the Monte Carlo simulation.

\subsection{Dissimilarity definition}

When building the hierarchy, the hierarchical level $H^i$ encodes information of the level below ($H^{i-1}$). This ensures that the information at the top-level accurately represents the information in the whole dataset. Therefore, for each subsequent hierarchy level, we transfer the manifold and structure relationships from bottom to top. For the first hierarchical level, the similarity between two data points $x_i$ and $x_j$ is computed using Equation~\ref{eq:kernel} based on euclidean distance ($d(x_i, x_j)$), as in Fig.~\ref{fig:overview-humap} (A). For higher levels, the similarities among landmarks in $H^i$ follow the data organization of $H^{i-1}$. Such similarities are computed based on the $k$-nearest neighbor graph of $H^{i-1}$.

On level $i$, for two landmarks, $l^i_u$ and $l^i_v$, we compute their similarity using two components: the intersection of the global and local neighborhoods. Fig.~\ref{fig:overview-humap} (D-F) illustrates these two components for two landmarks in green. The local neighborhood (in pink) consists of the k-nearest neighbors as usual, while the global neighborhood (computed only for $H^i, i \leq 1$) is found through random walks (in blue). Two landmarks that share more local neighbors are very close in the high-dimensional space since there are single data points that connect them (for landmarks $(l^i_u)$ and $(l^i_v)$, there is a data point $(l^i_p)$ in the neighborhood of $(l^i_u)$ and $(l^i_v)$). Landmarks sharing global neighbors have global relationships since there is a path between these landmarks. In Fig.~\ref{fig:overview-humap}, the landmarks in green only share a global relationship, highlighted in gray between steps (E) and (F).

We use random walks to compute the global neighborhood (the blue relationship in Fig.~\ref{fig:overview-humap}) similarly to the HSNE~\cite{Pezzotti2016} approach. In this case, we start $\omega$ random walks of length $\upsilon$ from each non-landmark ($x_k^{i-1}$) on the $k$-nearest neighbor graph of $H^{i-1}$. Then, when a random walk reaches a landmark, $(l^i_u)$, we add the non-landmark ($x_k^{i-1}$) to the ``representation neighborhood'' of landmark $l^i_u$. Notice that such a representation neighborhood (RNH) is created only for similarity computation. 

The local neighborhood component adapts our approach to preserve more of the local structures (see the pink relationship in Fig.~\ref{fig:overview-humap}). To this end, we use the $k$-nearest neighbors (NH) of each landmark $l^i_u$ to augment their representation neighborhoods with the $\beta\times |NH(l^i_u)|$ first nearest neighbors of $l^i_u$, where $\beta$ is a threshold between $0$ and $1$. Thus, the greater $\beta$, the more important the $k$-nearest neighbors, and the similarity measure will encode more local relations. HUMAP supports parameter tuning to perform better on neighborhood preservation. The similarity between two landmarks consists of the intersection between their global and local neighborhoods. The local neighborhood for a landmark is in the set of its nearest neighbors, so one can increase it by adjusting the parameter $\beta \in [0, 1]$ that adds $\beta|NH(i)|$ neighbors to the local neighborhood---by default, $\beta$ is set to $0$. We evaluated HUMAP with different values of $\beta$ and found a clear relationship between local neighborhoods and neighborhood preservation.
%
%

Finding these two components results in the representation neighborhood (RNH) for each landmark $l^i_u$ on level $H^i$. Such a neighborhood consists of a sparse matrix $RNH$ of dimensions $|H^i|\times|H^{i-1}|$ where $RNH_{u,v}$ is $1$ if $x_v^{i-1} \in$ RNH($l_u^i$) or $0$, otherwise. The similarity among landmarks is computed based on the intersection of representation neighborhoods, that is, the shaded area of Fig.~\ref{fig:overview-humap} (E-F). We compute the intersection with the $RNH$ matrices as follows:

\begin{equation}
\begin{split}
    \label{eq:similarity-relation}
    d_{RNH}(l^i_u, l^i_v) = \frac{(RNH_u)^TRNH_v}{M},
    \\M = max_m(\sum\limits_{n=1}^{|L^{i-1}|}RNH_{m,n})
\end{split}
\end{equation}

where $M$ is a normalization factor that ensures similarity between $0$ and $1$---notice that since $RNH$ is sparse, computing the above matrix multiplication is fast in practice. Note that the intersection between neighborhoods is also used in HSNE to compute the similarity between two landmarks. However, unlike in HSNE, we do not directly employ such a similarity measure to compute the embeddings. Instead, HUMAP extends the neighborhood to encode global information and uses UMAP's kernel to maintain its properties throughout the hierarchy.

The output of the above operation is a similarity measure, meaning that a value equal to 1 corresponds to landmarks sharing the same representation neighborhood, as illustrated by the operation (F) in Fig.~\ref{fig:overview-humap}. Thus, we transform it into a dissimilarity measure by subtracting the similarity value from 1. To benefit from UMAP's kernel function (Equation~\ref{eq:kernel}), this process of creating the neighborhood using Markov Chain also eliminates the necessity to perform another kNN step. Using the dissimilarity, we can simply apply the same kernel function on every hierarchical level.
%

\subsection{Generating the embedding}

Having the dissimilarity values ($1 - D_{RNH}$), where $D_{RNH}$ is a matrix with $d_{RNH}$ as rows, we are able to use them in Equation~\ref{eq:kernel} to generate the matrix $p_{ij}$ for each hierarchical level. Then, HUMAP, as in UMAP~\cite{McInnes2018} technique, applies the following matrix symmetrization to produce an undirected weighted graph:
%

\begin{equation}
    \label{eq:matrix-condition}
    p_{ij} = p_{i|j} + p_{p|j} - p_{i|j}p_{j|i}.
\end{equation}

The matrix $p_{ij}$ is used to produce an embedding that converges to positions in a low-dimensional space. While we refer to McInnes et al.~\cite{McInnes2018} for a detailed description of the UMAP algorithm, it is important to mention that an initial low-dimensional representation of the dataset is created using Spectral Embedding~\cite{Ng2001}. Then, $p_{ij}$ is used to reposition the data points using a force-directed graph layout whose convergence is performed by decreasing attractive and repulsive forces (given by $p_{ij}$) using Stochastic Gradient Descent~\cite{Ruder2016}.

\subsubsection{Subset projection}

Each landmark represents a set of data points on the level below when hierarchical levels are defined. It is crucial to specify the functionality of interaction with hierarchical levels and request more detail (more data points) of a data subset in order to keep track of such representation.

Each non-landmark data point on level $H^{i-1}$  is linked up with a landmark on level $H^i$ in two steps. The first step entails going through the landmarks iteratively and designating which of their neighbors will be influenced by them. A landmark will eventually attempt to "represent" a data point that has already been represented. As a result, we designate the data point as being a part of the area surrounding the closest landmark. The majority of data points are assigned to a landmark by this process, but it is still necessary to look for instances where data points are not in any landmark's neighborhood.
In this case, we iterate over each data point on $H^{i-1}$ to make sure that it is not associated with a landmark in case we want to search its neighborhood. This procedure is described in detail in Algorithm~\ref{alg:associate-landmark}. We examine the neighborhood of each non-landmark (Lines 2 to 5). If a neighbor, $v$, is a landmark (Line 6), we set $v$ to represent the current data point (Lines 7 and 8). If $v$ is not a landmark but is associated with one (Line 9), the landmark that represents $v$ will also represent the current data point (Lines 10 and 11).
Finally, we begin a depth-first search in the $k$-nearest neighbor graph and associate the first landmark found with the current data point if neither of these scenarios holds true for all neighborhoods.

\begin{algorithm}[]
\small
\caption{Associate Landmarks.}
\label{alg:associate-landmark}
\begin{algorithmic}[1]

\Procedure{Associate\_landmark}{X}
    \For{$ u \in X$}
        \If{ not landmark(u)}
            \State found $\leftarrow$ false 
            \For{$v \in NH(u)$} 
                \If{landmark(v)} 
                    \State found $\leftarrow$ true
                    \State associate[u] $\leftarrow$ v
                \ElsIf{ associate[v] $\neq$ NULL}
                    \State found $\leftarrow$  true
                    \State associate[u] $\leftarrow$ associate[v]
                \EndIf 
                
            \EndFor

            \If{not found}
                \State landmark $\leftarrow$  depth-first search on graph $X$ starting in $u$
                \State associate[u] $\leftarrow$ landmark
            \EndIf
            
        \EndIf
        
    \EndFor

\EndProcedure
\end{algorithmic}
\end{algorithm}

With the data points assigned to each landmark, users can select a list of indices on level $H^i$ to embed the associated data points on $H^{i-1}$. Thus, supposing the list of indices in $H^i$ is denoted by $I$, we only have to create the matrix $p_{mn}^{i-1} = \lbrace p_{mn}~ |~ A[m] \in I \wedge A[n] \in I\rbrace$, where $A[u]$ returns the landmark that represents the data point $x_u$.

\subsubsection{Mental map preservation}
\label{sec:mental-map}

\wm{When updates are made on the current dataset at time $t$, the concept of mental map preservation entails maintaining the same overall layout shape and point positioning~\cite{Archambault2011AnimationSM, Fujiwara2019}---these updates may involve new data points or modifications to the features of existing data points. In the case of interacting with hierarchical dimensionality reduction results, the updates correspond to new data points from lower hierarchy levels being projected.}

\begin{figure}[h!]
 \centering 
 \includegraphics[width=0.7\linewidth]{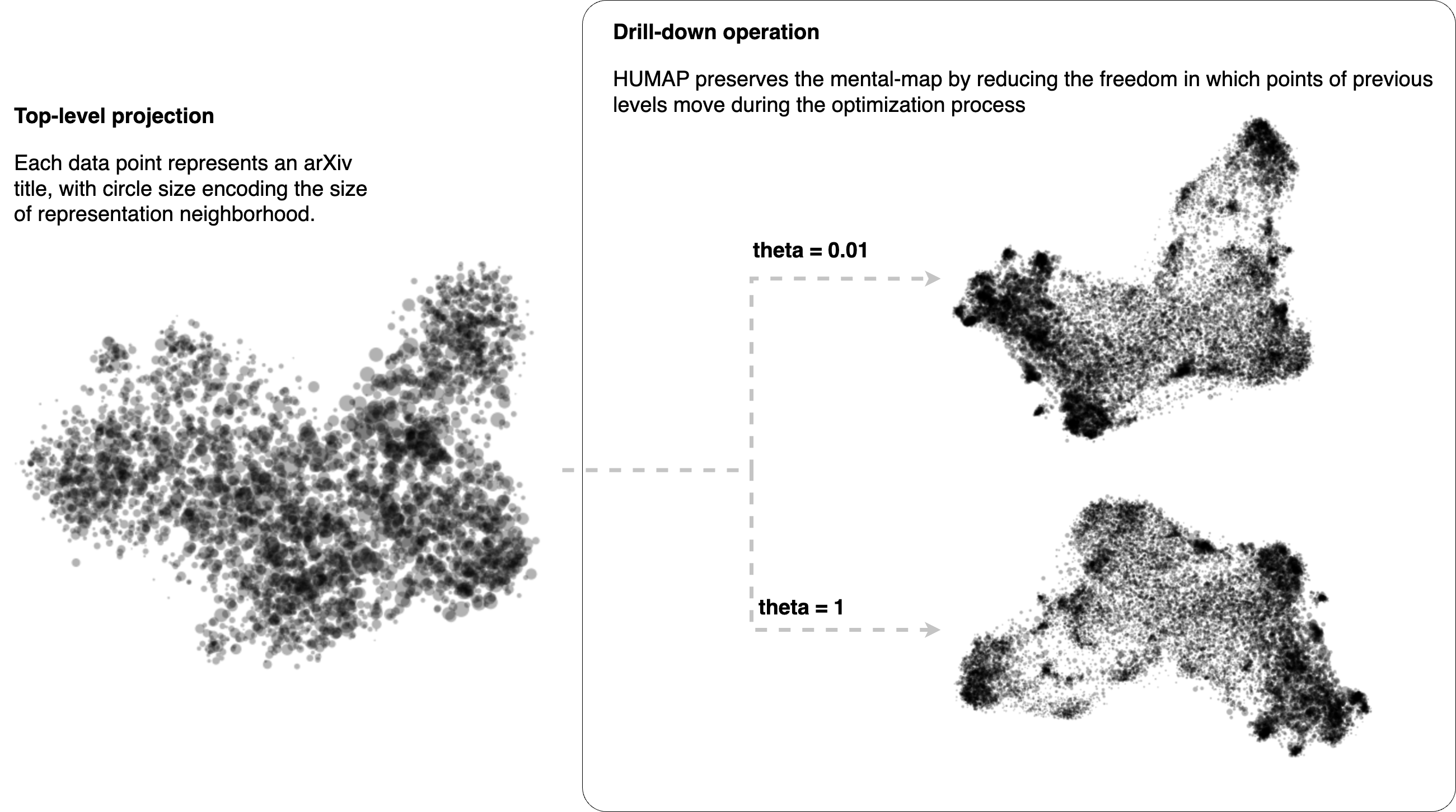}
 \caption{Hierarchical exploration with HUMAP using mental map.}
 \label{fig:mental-map}
\end{figure}

The subsets of data chosen for fine-grained exploration on level $l-1$ during hierarchical exploration must resemble the organization of level $l$, especially for unlabeled datasets. Rauber et al.~\cite{Rauber2016} and Fujiwara et al.~\cite{Rauber2016} also present strategies to deal with mental map preservation in the context of dimensionality reduction, even though the majority of the work on mental map preservation is limited to dynamic graph drawing~\cite{Xu2012, Leydesdorff2008}. In both situations, the goal is to ease the analysis burden during exploration and facilitate smoother embedding layout changes. Fig.~\ref{fig:mental-map} shows successive hierarchical projections using HUMAP, whether or not the mental map is preserved. Note that the embedding generated with no mental-map preservation (i.e., $\theta = 1$) might deceive the user in thinking that the overall organization of the layout was preserved.

HUMAP uses the coordinates of higher---and already projected---data points on level $l$ to guide the positioning of level $l$-1. When initializing the low-dimensional representation before Stochastic Gradient Descent (SGD) optimization, the coordinates for the projected data points in $l$ are used as starting points together with the remaining data points initialized with Spectral Embedding---these coordinates only move a fraction during SGD optimization. Thus, we also provide this fraction hyperparameter $\theta$ to be tuned according to one's needs, although we empirically find that $\theta = 0.01$ yields embeddings that best preserve the mental map throughout the hierarchy expansion.

\begin{figure*}[!htb]
 \centering 
 \includegraphics[width=1\textwidth]{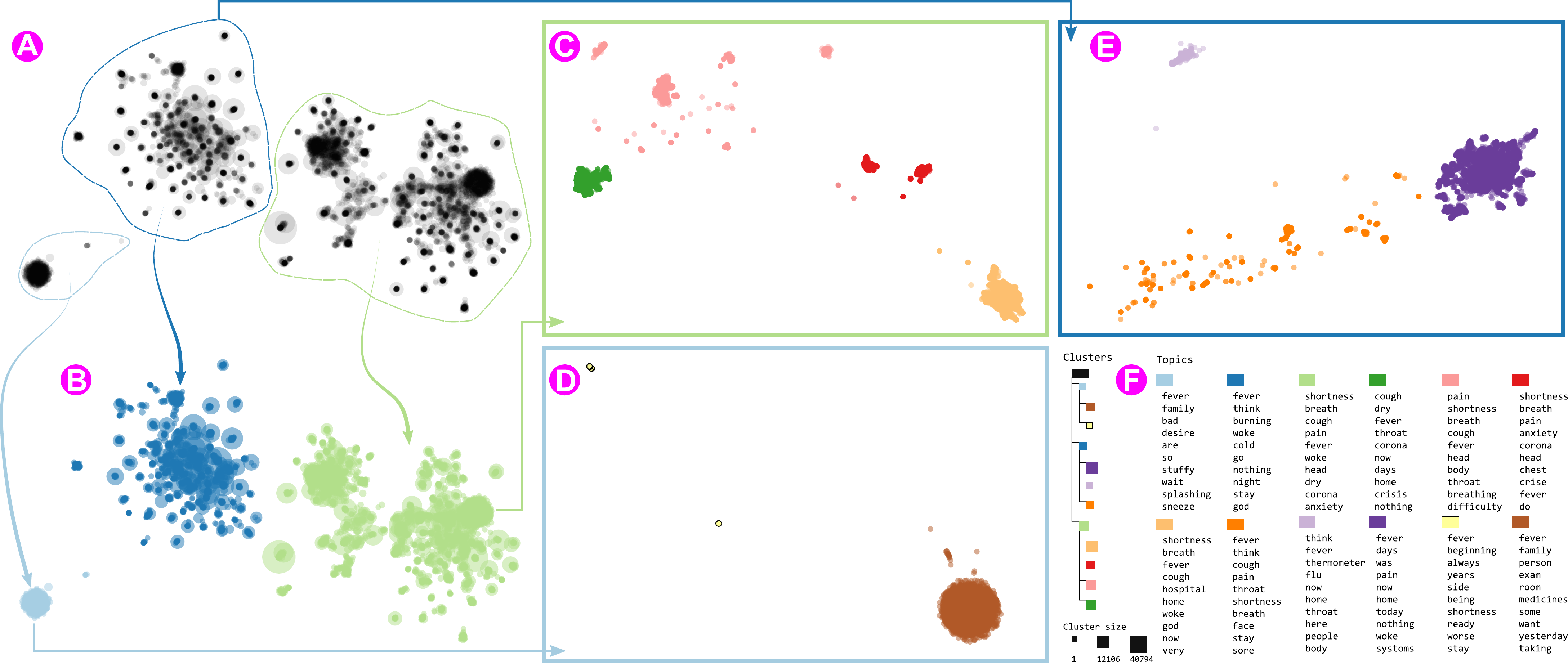}
 \caption{HUMAP exploration and annotation of a document collection of COVID-19 \textit{tweets}. The top-level hierarchy level shows unlabeled data points and three major structures (A). We annotate these three clusters (B) and compute their topics computed (F). For each cluster in (B), we also project their corresponding level (and final) hierarchy to look for other patterns, annotating the dataset (C, D, E) and computing their topics.}
 \label{fig:covid19_analysis}
\end{figure*}

\section{Case study}
\label{sec:case-studies}

This section aims to investigate the relationship of \textit{tweets} about COVID-19 symptoms in the São Paulo state, Brazil~\cite{MarcilioJr2021_Contrastive}. We use HUMAP to explore it using two hierarchical levels, aiming at discovering dominant structures and detailed information about these dominant structures through interaction. The dataset was scraped from Twitter$^{\tiny{\textregistered}}$ by querying COVID-19 symptoms (fever, high fever, cough, dry cough, difficulty breathing, and shortness of breath) in the territory of São Paulo state (Brazil) from March 2020 to August 2020. The authors classified the \textit{tweets} according to their relevancy (relevant or not relevant) to COVID-19 infection. For this case study, we set HUMAP to freely find the embeddings as we drill-down the hierarchy by not projecting the low-levels bases on the higher levels. 

To explore the dataset, we manually defined clusters in the visual space using lasso selection, as shown in Fig.~\ref{fig:covid19_analysis}(A-B). After associating each data point to a cluster using this procedure, we compute topics (F) and proceed to the second and lowest level of the hierarchy, choosing the desired cluster (e.g., cluster~\img{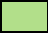}). Then, we also compute the topics for the manually defined cluster of the new hierarchical level.

Fig.~\ref{fig:covid19_analysis}(B) shows three clusters with different characteristics. First, cluster \img{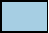} is very cohesive and dissimilar from the other two dominant structures. Second, due to visual proximity, cluster \img{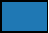} and \img{figures/light-green} share some information but contain various substructures that need further investigation. The topics for these three major clusters further explain their organization in the visual space (F). That is, cluster \img{figures/light-blue} presents important terms related to respiratory problems, such as \texttt{stuffy} nose, \texttt{sneeze}, or \texttt{splashing}. Other important terms in this cluster can indicate \textit{tweets} about individuals waiting (\texttt{wait}) for COVID-19 \texttt{exams} or are associated with \texttt{desire} to \texttt{sneeze}, which supports a hypothesis that this cluster corresponds to individuals worrying about symptoms. Cluster \img{figures/dark-blue} shows three terms in the topic associated with fever: \texttt{fever}, \texttt{burning}, and \texttt{cold}. Other terms such as \texttt{think}, \texttt{woke}, and \texttt{night} might be associated to phrases describing experiences with COVID-19 symptoms, such as: ``I \textbf{think} I have \textbf{fever}'', ``Today I \textbf{woke-up} in the middle of the \textbf{night} \textbf{burning} in \textbf{fever}''. Lastly, cluster \img{figures/light-green} is associated with dry cough and shortness of breath. The terms for this cluster in Fig.~\ref{fig:covid19_analysis}(F) show that the \textit{tweets} talk about these symptoms while the term \texttt{anxiety} could be the cause of shortness of breath~\cite{Smirni2020}.

We proceed to cluster \img{figures/light-green} to investigate its substructures and retrieve more information about the \textit{tweets} associated with it. Then, we define the clusters and compute the topics, as shown Fig.~\ref{fig:covid19_analysis} (C) and (F). Here, there are a few interesting patterns. The first one is that the previous hierarchy level sufficiently gives an overview of the data organization since the topic for cluster \img{figures/light-green} encodes most of the information expressed in these sub-clusters. The second and most interesting aspect is the global relationship among these structures apparent in the embedding. The topics retrieved from each local structure explains this aspect. The leftmost cluster (\img{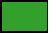}) shows terms related to \texttt{dry cough}, \texttt{throat} pain, and a few other important terms. Cluster \img{figures/dark-green} has a relationship with cluster \img{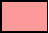}, which also adds \texttt{difficult breathing}, \texttt{body pain}, and headache (``pain in the \textbf{head}'', using a direct translation from Portuguese). In cluster \img{figures/light-red}, \texttt{shortness of breath}, \texttt{anxiety}, \texttt{crisis}, and breast become more important. These symptoms might be easily confused with anxiety crisis, a common problem during COVID-19 pandemic~\cite{DeBoni2020}. The last cluster (\img{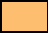}) is the most related to COVID-19 symptoms, showing the majority of term: \texttt{shortness of breath}, \texttt{fever}, and \texttt{cough}.

Analyzing the substructures of cluster \img{figures/dark-blue}, we define three clusters as shown in Fig.~\ref{fig:covid19_analysis} (E). As well-defined by the higher-level cluster, all data points refer mainly to the COVID-19 symptom of fever. However, there are a few characteristics that might explain the differentiation of these clusters. For example, cluster \img{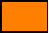} present terms related to \texttt{cough}, \texttt{shortness} of \texttt{breath}, which do not appear in the other clusters. Lastly, cluster \img{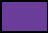} has an interesting characteristic since its topic suggest that individuals are commenting about the period in which they experience fever: \texttt{fever}, \texttt{days}, \texttt{was}, \texttt{today}, and \texttt{home}. The most cohesive cluster of top-level (\img{figures/light-blue}) and its two subclusters (\img{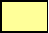} and \img{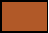}) revealed by drilling-down the hierarchy show tweets that seem to talk more about daily aspects than truly COVID-19 symptoms.

\wm{Now, to make readers more informed about differences between traditional and hierarchical approaches, we present an analysis of the same dataset using UMAP.}

\subsection{Comparison to UMAP}

\begin{figure}[tb]
 \centering 
 \includegraphics[width=\linewidth]{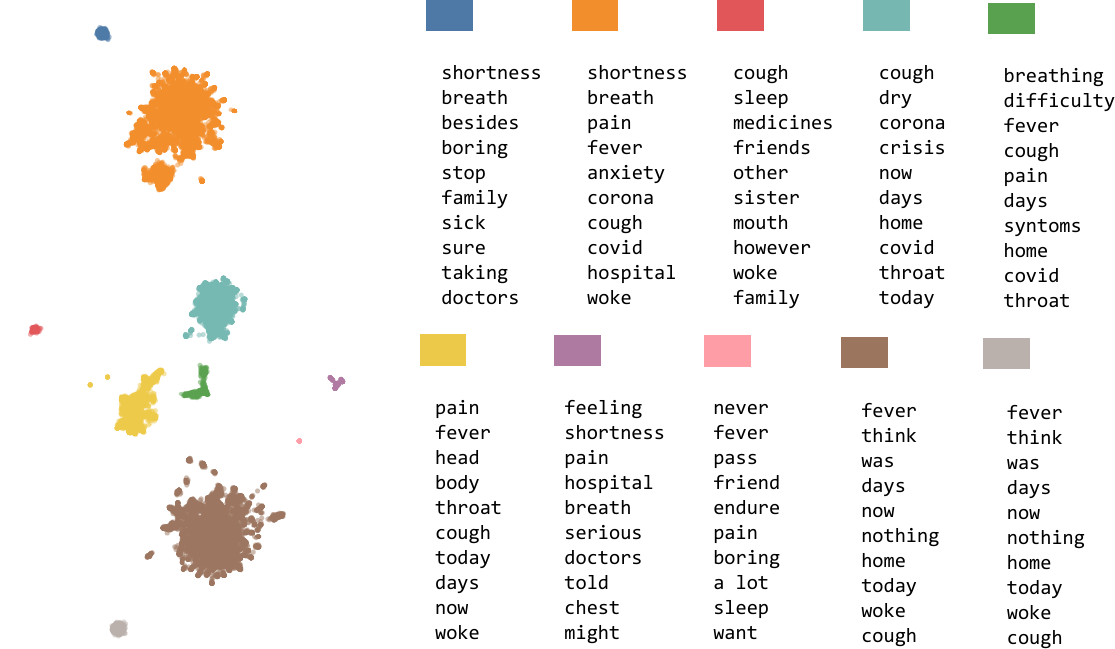}
 \caption{Manually annotated UMAP projection with cluster topics.}
 \label{fig:umap-exploration}
\end{figure}

\wm{Fig.~\ref{fig:umap-exploration} shows and annotated UMAP projection of the same dataset in an intent to showcase the differences between analysis using hierarchical and one-level dimensionality reduction techniques.}

\wm{The first thing to notice is that the layout generated by UMAP is somewhat similar to HUMAP's highest level in Fig.~\ref{fig:covid19_analysis}. Second, the overall structure of topics shortness of breath, cough, and fever is also seen and dictates the relationship among clusters---from the top (blue cluster) to the bottom (gray cluster).}

\wm{Hierarchical approaches allows to discover different relationship that would require more expertize and time to tune hyperparameters if traditional approaches were employed. Comparing the two types of analysis, hierarchical exploration enables identifying fine grained information containing in datasets.}

\begin{figure*}[!t]
 \centering 
 \includegraphics[width=\textwidth]{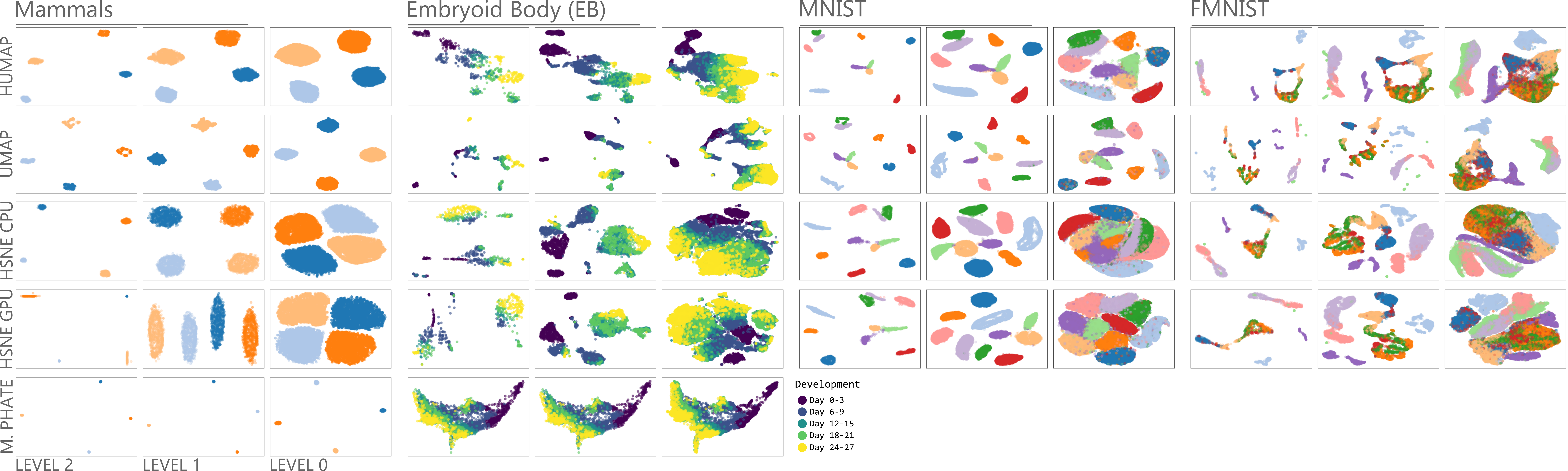}
 \caption{Visual analysis of the embeddings generated for top and lowest hierarchical levels using a three-level hierarchy. For each dataset, top-level embedding appears on the left, and the lowest level (whole dataset) appears on the right.}
 \label{fig:analysis}
\end{figure*}


\section{Numerical evaluation}
\label{sec:experiments}

In this section, we numerically evaluate HUMAP\footnote{https://github.com/wilsonjr/humap} and compare it against existing HDR techniques: Hierarchical Stochastic Neighbor Embedding (HSNE)\footnote{https://github.com/biovault/nptsne}~\cite{Pezzotti2016}, Multiscale PHATE\footnote{https://github.com/KrishnaswamyLab/Multiscale\_PHATE}~\cite{Kuchroo2020}, and UMAP\footnote{https://github.com/lmcinnes/umap}~\cite{McInnes2018}. We cannot control the number of data points in each hierarchy level in HSNE and Multiscale PHATE. Thus, we generated HSNE with three levels and fit a HUMAP hierarchy with three levels based on the number of data points in each HSNE level. Multiscale PHATE does not accept a parameter to specify the number of levels. After fitting the Multiscale PHATE hierarchy, we searched for a level with a size similar to the top-level produced by HUMAP and HSNE techniques. Finally, we evaluate UMAP to demonstrate its differences from HUMAP regarding mental map and structure preservation by projecting the same data points generated in the HUMAP hierarchy.
%

We use the following datasets for evaluation. MNIST~\cite{LeCun2010} is a dataset composed of $70,000$ $28\times 28$ pixel grayscale images of handwritten digits classified into ten classes (from $0$ to $9$), where each flattened image results in a $784$ dimensional vector. Fashion MNIST~\cite{Xiao2017} (FMNIST) is a dataset composed of $70,000$ $28\times 28$ pixel grayscale images of fashion items (clothing, footwear, and bags) divided into ten classes; each flattened image results in a $784$ dimensional vector. Mammals is a synthetic dataset designed to have four well-separated classes, and it consists of $20,000$ data points described by $72$ dimensions. Embryoid Body (EB) is a single-cell RNA sequencing dataset for embryoid body data generated over $27$ days~\cite{Moon2018} divided into five periods. For this dataset, we aim to visualize the development of the $31,000$ cells after preprocessing them and using the first 50 principal components to perform the projections. Note that, FMNIST, MNIST, and Mammals datasets are meant to produce cohesive clusters after embedding, while the ideal result for Embryoid Body is to provide an understanding of its continuous structures---this is an attempt to understand how HUMAP compares to different techniques regarding their most known characteristics, that is, to cluster data (HSNE, derived from SNE) and to emphasize continuity (Multiscale PHATE derived from PHATE). The experiments were performed in a computer with the following configuration: Intel(R) Core(TM) i7-8700 CPU @ 3.20 GHz, 32 GB RAM, Ubuntu 64 bits, NVIDIA GeForce GTX 1660 Ti 22 GB.

\begin{table}[!htb]
\centering
\begin{tabular}{@{}lrr@{}}
\toprule
Dataset             & \multicolumn{1}{l}{Size} & \multicolumn{1}{l}{Dimensions} \\ \midrule
Mammals             & 20000                    & 72                       \\
Embryoid Body (EB)  & 31000                    & 50                     \\
MNIST               & 70000                    & 784                      \\
FMNIST              & 70000                    & 784                      \\ \bottomrule
\end{tabular}
\caption{Datasets used for experimentation.}
\label{tab:computing-time}
\end{table}

Fig.~\ref{fig:analysis} depicts the embeddings for the hierarchy levels. \wm{HSNE shows a good relationship between clusters in the top-level embedding, but fails to maintain that structure on consecutive embeddings (levels 1 and 0).} The continuous nature of the biological data is not shown for the EB dataset. One significant problem, to sum up, is that the mental map cannot be maintained throughout the hierarchy.


Due to their high dimensionality, Multiscale PHATE was unable to produce embeddings for the FMNIST and MNIST datasets. Despite revealing the continuous nature of the EB dataset, it produced embeddings for the mammals dataset that made analysis challenging because the data points for each cluster were too close to one another. On the other hand, HUMAP maintains the mental map through all levels of the hierarchy by encoding the data from lower levels at higher levels. \wm{The main challenge for Multiscale PHATE is scaling with respect to number of dimensions.}


Finally, UMAP only reveals the correct overview for the simpler datasets on level 2. (e.g., \textit{mammals} and \textit{MNIST}). UMAP is unable to uncover the structures displayed at the lowest level for the \textit{EB} and \textit{FMNIST} datasets. This outcome was anticipated because UMAP's primary objective is not to transfer the relationship between data points to the projection sample. The ability to successfully encode datasets with less data is achieved by HUMAP, which projects higher levels based on relationships at lower levels.

\begin{figure}[tb]
 \centering 
 \includegraphics[width=1\linewidth]{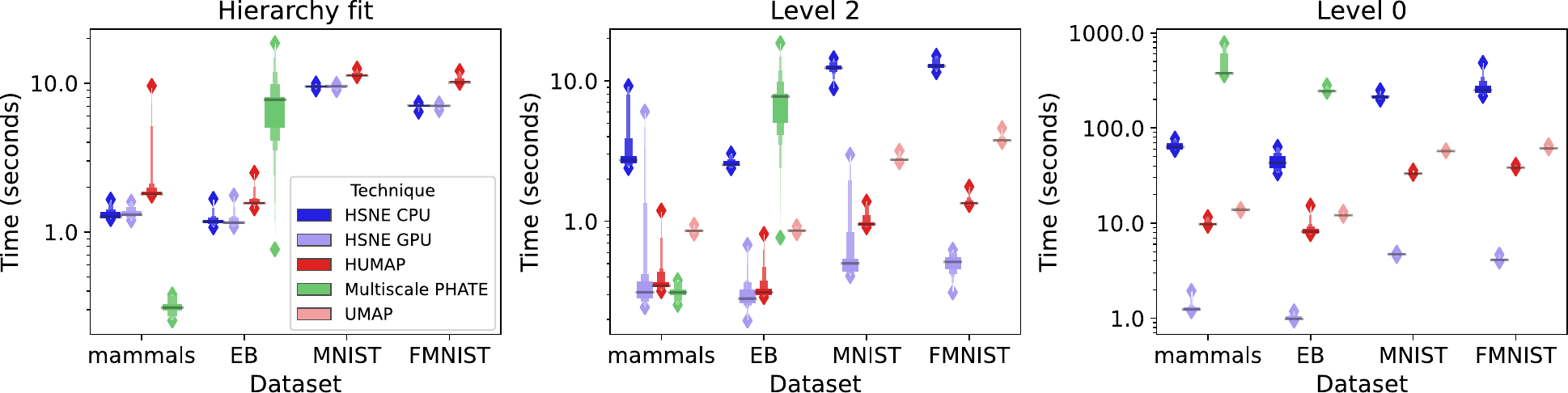}
 \caption{Runtime execution in seconds using a log$_{10}$ scale. The boxen-plots show the runtime for $20$ runs of each technique to fit the hierarchy (Hierarchy fit), to embed the top level (Level 2), and to embed the whole dataset (Level 0).}
 \label{fig:analysis-time}
\end{figure}

For the numerical evaluation, we compare the techniques using well-known and traditional quality metrics for embedding evaluation (e.g., \textit{Trustworthiness} and \textit{Continuity} in \textbf{Supplementary File}) as well as the metrics that appear in the related works, such as \textit{DeMaP}~\cite{Moon2019} and \textit{Neighborhood Preservation}~\cite{Paulovich2008} that aim to evaluate how the low-dimensional embeddings represent the structures in higher dimensions.

\subsection{Running-time and Scalability}

\wm{ Fig.~\ref{fig:analysis-time} displays boxen-plots in log scale for runtime execution in seconds in order to fit the hierarchy and embed levels 2 and 0}. HUMAP offers comparable runtime execution to HSNE in GPU, with the exception of level 0, making it a promising strategy for users with limited resources when reasonable runtime execution is required. When fitting the hierarchy, HUMAP takes slightly longer than HSNE to run on the CPU, but it is quicker when embedding the hierarchy levels. When many data points are embedded, Multiscale PHATE seems unreasonable for interactive applications (for example, level 0 of mammals and EB datasets). \wm{Finally, the difference between HUMAP and UMAP narrows as we approach to the size of the whole dataset.}


\subsection{Neighborhood Preservation}

\begin{figure}[htb]
 \centering 
 \includegraphics[width=\columnwidth]{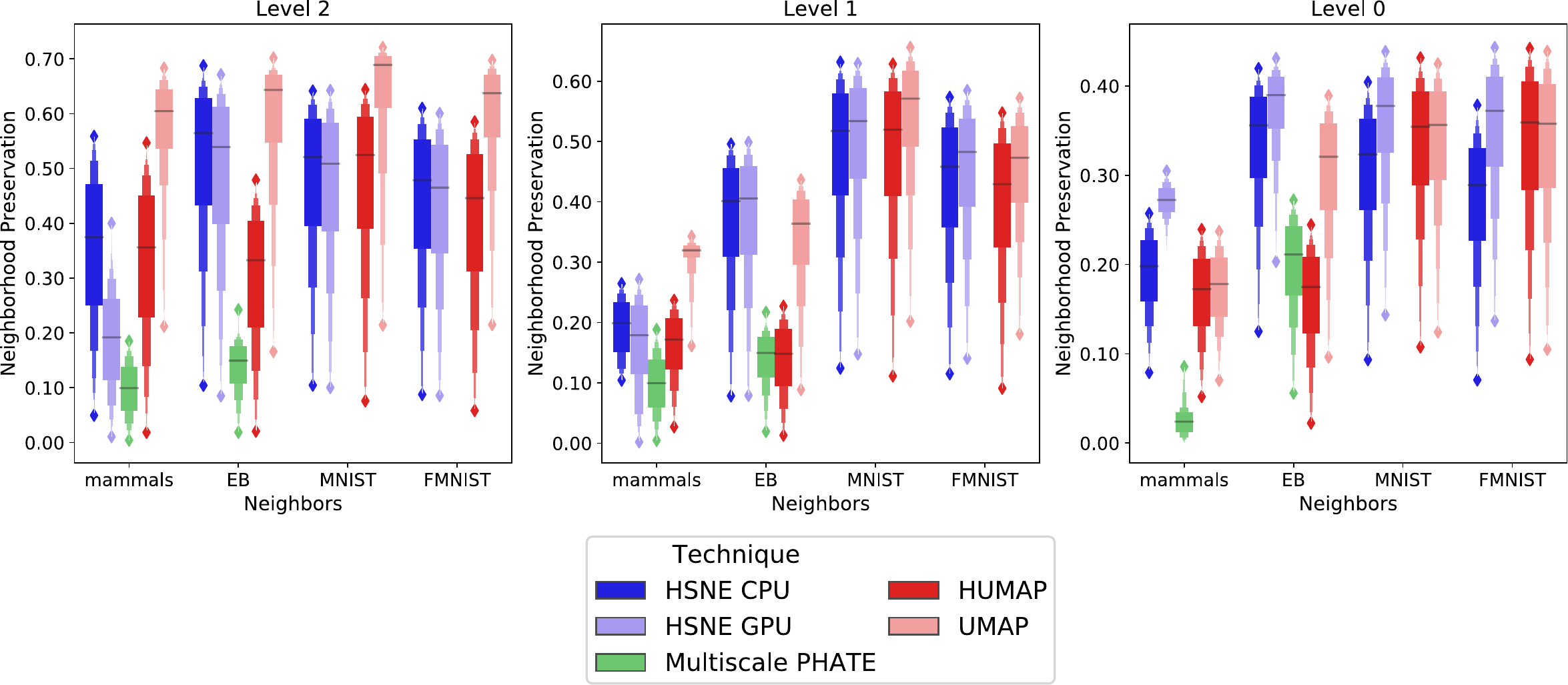}
 \caption{Neighborhood Preservation after embedding on $\mathbb{R}^2$. HSNE outperforms HUMAP by a great margin only for \textit{EB} dataset.}
 \label{fig:analysis-np}
\end{figure}

Fig.~\ref{fig:analysis-np} depicts the neighborhood preservation (NP) for a variable number of neighbors, $k$ ($k \in [1,30]$). Such a metric calculates the mean ratio of neighbors preserved in the projection. 
%

The \textit{mammals} dataset is intended to have clearly defined classes, whereas the \textit{EB} dataset is intended for continuity analysis. Because it does not just concentrate on local structures when projecting such datasets, HUMAP achieves a lower NP. The characteristics of SNE are inherited by HSNE, which exhibits a higher NP for cluster separation (see the EB projections, for example).

We evaluated HUMAP using various values of $\beta$ (which controls the local neighbors' contribution to the similarity among landmarks), and the results demonstrated the link between local neighborhoods and neighborhood preservation. In addition, for the MNIST dataset at the top hierarchy levels, HUMAP outperforms HSNE by a considerable margin with the addition of only $3\%$ of the current neighborhood (100 neighbors). Similar data points are clustered by HUMAP as a result of the $\beta$ parameter; interested readers can visualize the analyses and projections for this scenario in the \textbf{Suppl. File} (Figs. 3 and 4).

\subsection{DEMaP}

\begin{figure}[htb]
 \centering 
 \includegraphics[width=\columnwidth]{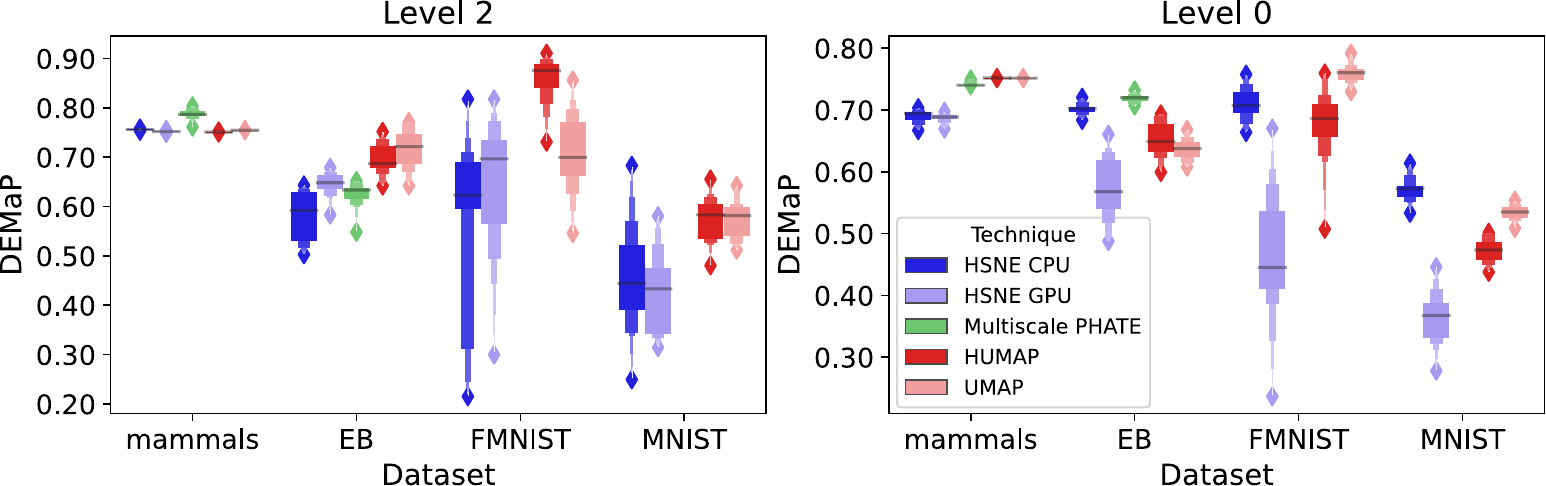}
 \caption{Evaluation of techniques' ability to represent complex structures such as clusters, manifolds, and other relationships. }
 \label{fig:analysis-demap}
\end{figure}

We employ the DEMaP metric~\cite{Moon2019}, which calculates the Spearman correlation between geodesic distances on high-dimensional and euclidean distances on low-dimensions, to assess how well the techniques convey manifolds, clusters, and other high-dimensional space structures. On level 2, HUMAP outperformed HSNE, UMAP, and GPU-based HSNE for the MNIST and FMNIST datasets, as shown in Fig.~\ref{fig:analysis-demap}. The pair of distributions (HUMAP, UMAP) for the EB, FMNIST datasets on level 2 and for the EB dataset on level 0 are not statistically different after a t-test, but the others are with a p-value of at least $<$ 0.00001 (see \textbf{Supplementary File} - Table 2 for the details). When the entire dataset is embedded for the mammals dataset, HUMAP displays higher values, providing proof that our method is stable across hierarchical levels. While HSNE and Multiscale PHATE produce better results at the top of the hierarchy, they omit key details as one descends hierarchy---the relationship among clusters is lost on the HSNE side. \wm{Finally, HUMAP conveys the continuous structures present in the EB dataset even for the top-level embedding while M. PHATE and HSNE CPU show higher DEMaP values at level 0.}

Fig.~\ref{fig:analysis} demonstrates the mental map preservation across hierarchical levels. In particular for unlabelled datasets, the preservation of the mental map is crucial to avoid misleading users and increasing cognitive load during exploration; this characteristic also holds for subsets of data points, which we address in the following section. According to the DEMaP metric, the fixing term for maintaining the mental map lowers the quality of the entire projection (level 0). We restrict the flexibility of the optimization algorithm when positioning the data points when we set the embedding of lower hierarchy levels to follow the pattern of higher hierarchy levels. In the \textbf{Supplementary File} (Section 5), we provide the same analysis by setting free the optimization algorithm ($\theta = 1$), which results in higher DEMaP values.

%

\subsection{Projecting subsets of data}

\begin{figure}[!htb]
     \centering
     \subfloat[][Top-level and the selection of subsets for detailed analysis.]{\includegraphics[width=\columnwidth]{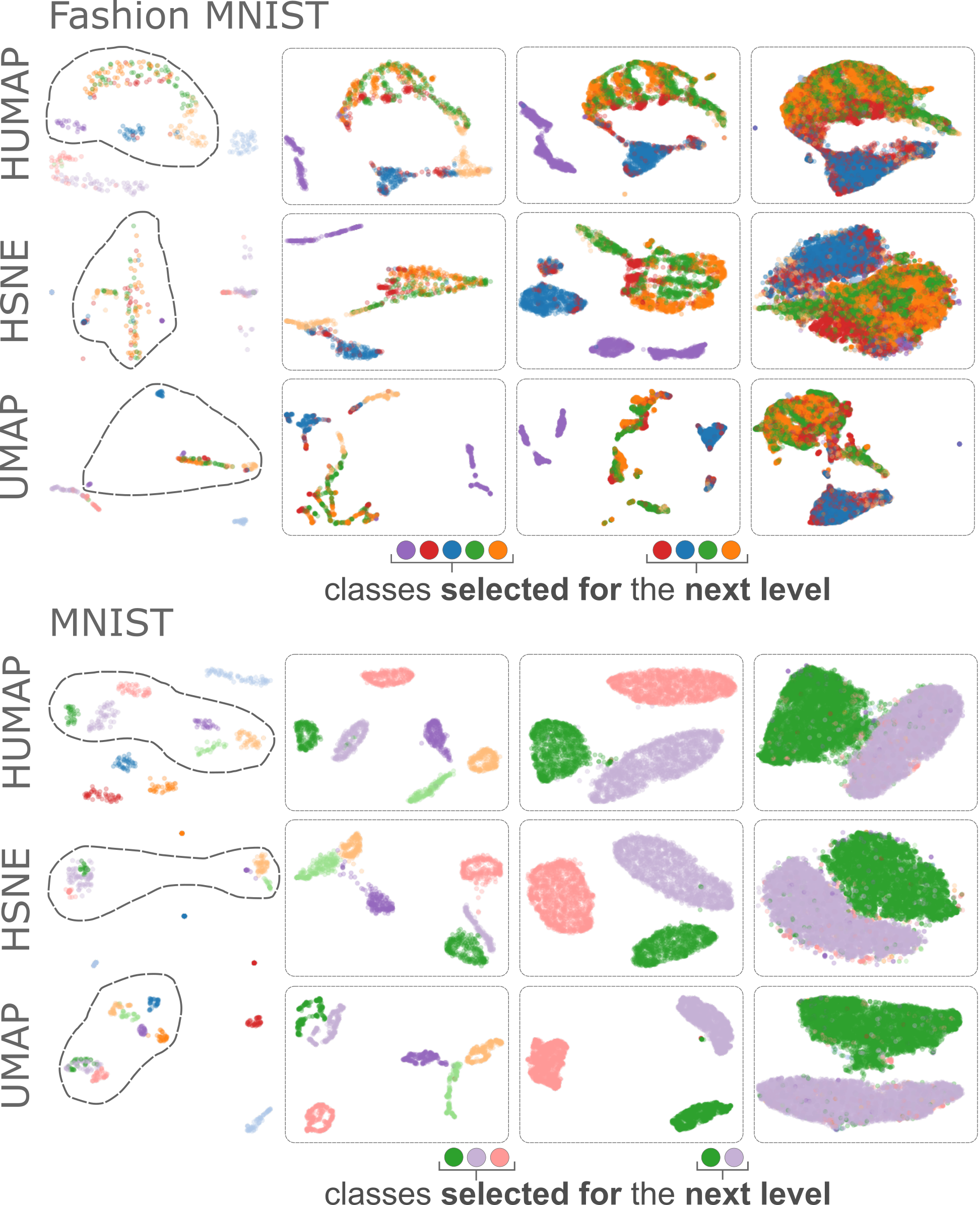}\label{fig:drill-down}}
     
     \subfloat[][DEMaP values to evaluate the quality of embeddings.]{\includegraphics[width=\columnwidth]{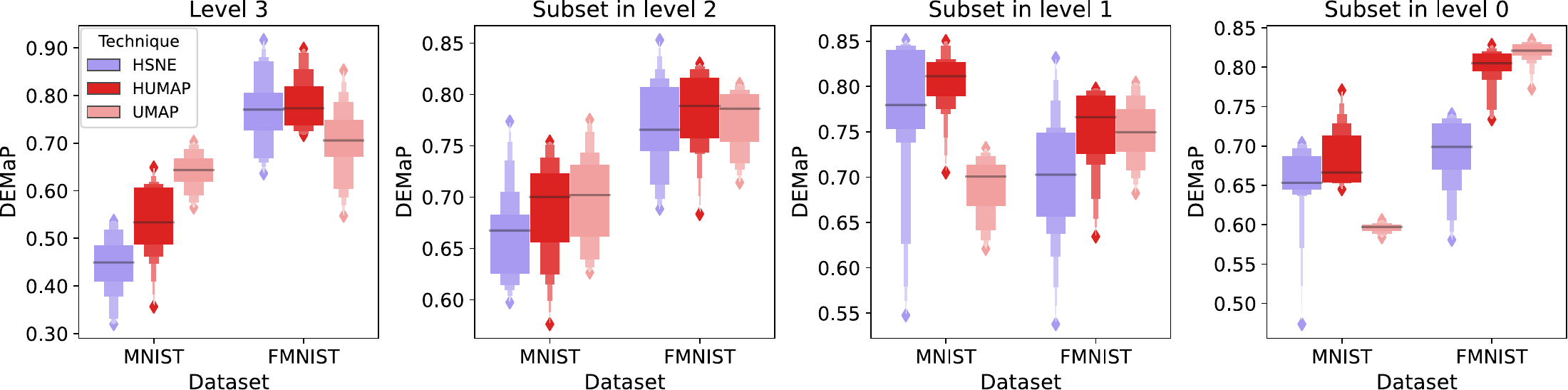}\label{fig:analysis-drill-down}}
     
     \caption{Using HUMAP, UMAP, and HSNE to project a subset of data.}
     \label{fig:evaluating-drill-down}
\end{figure}

In this section, the methods are compared according to how well they can project subsets of data points. Since there is no way to drill down a hierarchy based on classes of data points, we do not analyze Multiscale PHATE. Fig.~\ref{fig:drill-down} illustrates the same pattern as Fig.~\ref{fig:analysis}, in which HUMAP conveys top-level structures and maintains the mental map throughout the hierarchy. In this case, we project a four-level hierarchy, adding complexity to the analysis. In Fig.~\ref{fig:drill-down}, the first level's subset was lasso-selected, and the subsequent levels' subsets were chosen according to specific classes.

In all scenarios, as shown in Fig.~\ref{fig:analysis-drill-down}, HUMAP has a higher DEMaP than HSNE, and it only falls short of UMAP on the fourth level for the MNIST dataset—notice that UMAP by itself does not take the relationship between higher levels into account for projection. 

\subsection{Other quantitative evaluations}

After dimensionality reduction, additional quality metrics can be used to assess the final embedding. In order to broaden the analysis performed in this paper, another set of comparisons are presented in the \textbf{Supplementary File}. In the \textit{Runtime} analysis, we carefully examine how each HUMAP step is impacted by the dataset size and dimensionality as well as how it compares against the alternatives. We demonstrate in the \textit{Reproducibility} analysis that HUMAP can be just as effective as UMAP in terms of the reproducibility of subsequent runs. Finally, we also performed a parameter analysis to comprehend HUMAP's behavior for various parameter configurations.
\section{Discussion}
\label{sec:discussion}

The experiments demonstrate that HUMAP is competitive with GPU-based methods in terms of running time execution while outperforming current approaches in their ability to represent structures found in high-dimensional spaces. It also preserves the relationships among clusters and other structures at various levels of the hierarchy. Since HUMAP preserves the overall structures of higher hierarchy levels at lower hierarchy levels, it is valuable ideal tool for progressive analysis~\cite{Fujiwara2020}. Here, we go over a few of the features of our method.


\paragraph{\textbf{Reproducibility.}} At the expense of a longer run-time execution, HUMAP allows for the generation of reproducible embeddings. In other words, we need to disable a few parallel execution steps in the components for hierarchical definition and projection. When we measure the correlation of data points across successive projections~\cite{Becht2018}, experiments demonstrate perfect reproducibility for HUMAP for various datasets.

\paragraph{\textbf{Embedding initialization and CPU/GPU implementations.}} Recent research demonstrates that the initialization of the low-dimensional representation significantly affects the performance of these two algorithms, with PCA~\cite{Jolliffe1986} being preferable to random initialization for t-SNE~\cite{Kobak2021InitializationIC}. T-SNE with random initialization is used in HSNE. This fact may help to explain why the HSNE yielded lower results for DEMaP metric at the lowest level (the entire dataset). However, HSNE effectively encodes the global neighborhood at higher hierarchy levels, so we believe integrating this initialization step would still not impact the final results. The embedding cannot currently be started with PCA initialization using the official HSNE implementation. PCA initialization, however, does not address the issue of the mental map not being preserved across hierarchy levels.

The performance of HSNE on the CPU being consistently superior to HSNE on the GPU is another intriguing aspect of the numerical comparison. But why do we observe such a contrast? First, the embeddings are computed by HSNE on GPU using the GPGPU approximation. Although it produces acceptable embeddings, this approximation compromises accuracy for runtime. The Barnes-Hut implementation is used by HSNE on the CPU to compute the embeddings with $\theta$=0, the speed/accuracy trade-off parameter, which ranges from 0 (exact t-SNE) to 1 (approximated t-SNE).

\paragraph{\textbf{Landmark selection.}} We must choose representative data points (or landmarks) to be a subset of the dataset for a specific hierarchy level in order to define different hierarchy levels. For this task, we empirically assessed a number of strategies. Only evenly distributed datasets can benefit from choosing landmarks from denser neighborhoods; otherwise, hierarchy levels might only be able to encode some of the data (the dense regions). Finally, because of the computational complexity and difficulty in accurately capturing the data organization given outliers and other data structures, such as clusters of different shapes, clustering approaches also do not work well for our problem.

\paragraph{\textbf{HUMAP generalizability and scalability.}} In order to show the efficacy of HUMAP for hierarchical exploration, we compared it with various hierarchical dimensionality reduction methods, as well as with UMAP, throughout the experiments. Despite the fact that we are aware that different hyperparameters might produce varying results, we use the techniques' default parameters in this context to provide the analysis. We attempted to track various applications for a dimensionality reduction technique, from comprehending clusters for multidimensional datasets to visualizing single-cell sequencing data, although a more thorough study would be an interesting topic for future research activities. In the \textbf{Supplemental File} that is included, we also provide a comprehensive examination of HUMAP's scalability and generalizability for different hyperparameters.

\paragraph{\textbf{Applications.}} \wm{Another important aspect besides the technical differences among these techniques is related to where they can be employed. Obvious applications are in scenarios where one aims to understand hierachical differences within structures of data, such as structural differences in research papers~\cite{Gonzalez2023} and sub-cell types~\cite{Unen2017}. Finally, DR techniques have been extensively used to monitor the training of machine learning models, which makes hierarchical approaches suitable to summarize information.}

\section{Conclusion and Future Work}
\label{sec:conclusion}

For the analysis of high-dimensional data, DR techniques are excellent tools. Traditional methods, however, are unable to reveal substructures while giving a dataset's overview. Hierarchical DR approaches offer analysis that adheres to the mantra of visualization, in which analysts concentrate on crucial information and retain details as needed. However, the hierarchical approaches described in the literature either cannot be applied to a wide range of dataset types or do not preserve the mental map throughout the hierarchy levels.
%

\wm{We introduced HUMAP, a novel hierarchical DR technique that provides a viable alternative to high-dimensional data analysis and includes tunable parameters that make it simple to focus on global or local neighborhood preservation while also maintaining the mental map.}

Future research will expand on our technique to examine distance distance metrics between landmarks and data points. Additionally, we intend to implement GPU versions to enable users to delve deeper into HUMAP's capabilities. Finally, we intend to conduct user experiments on the significance of mental map preservation for hierarchical approaches and investigate novel strategies to evaluate hierarchical DR techniques.

\acknowledgments{
The authors wish to express their gratitude to all the reviewers who took the time to review this paper. Your insightful comments and feedback were invaluable and significantly contributed to enhancing the quality and clarity of the manuscript. This work was supported by Fundação de Amparo à Pesquisa (FAPESP) [grant number \#2018/17881-3] and the Coordenação de Aperfeiçoamento de Pessoal de Nível Superior (CAPES) [grant number \#88887.487331/2020-00].
}

\bibliographystyle{abbrv-doi}

\bibliography{template}

\end{document}